\pdfoutput=1
\documentclass[11pt,a4paper]{article}

\usepackage[dvipsnames]{xcolor}
\usepackage[hyperref]{acl2021}

\usepackage[utf8]{inputenc}

\usepackage{times}
\usepackage{latexsym}
\usepackage{xspace}
\usepackage{enumitem}

\usepackage{url}
\usepackage[dvipsnames]{xcolor}
\usepackage{booktabs}
\usepackage{diagbox}
\usepackage{multirow}
\usepackage{balance}
\usepackage{amsfonts}
\usepackage{amsmath}
\usepackage{amssymb}
\usepackage{graphicx}
\usepackage{subfigure}
\usepackage{microtype}
\usepackage{wrapfig,lipsum}
\usepackage{algorithm}
\usepackage{algorithmicx}
\usepackage{algpseudocode}
\usepackage{makecell}
\usepackage[T1]{fontenc}
\usepackage{soul}
\usepackage{tabularx, booktabs}
\usepackage{todonotes}
\usepackage{comment}
\newcommand{\cbb}{\mathbf{c}}
\newcolumntype{Y}{>{\centering\arraybackslash}X}
\newcolumntype{L}{>{\arraybackslash}X}
\newcolumntype{M}[1]{>{\centering\arraybackslash}m{#1}}

\algnewcommand\algorithmicinput{\textbf{Input:}}
\algnewcommand\algorithmicoutput{\textbf{Output:}}
\algnewcommand\algorithmicparameter{\textbf{Parameters:}}
\algnewcommand\INPUT{\item[\algorithmicinput]}
\algnewcommand\OUTPUT{\item[\algorithmicoutput]}
\algnewcommand\PARAMETER{\item[\algorithmicparameter]}

\newcommand{\clstoken}{\texttt{[CLS]}}
\newcommand{\septoken}{\texttt{[SEP]}}

\newcommand{\emptyspan}{\textsc{null}}

\newcommand{\eat}[1]{\ignorespaces}
\input{Definitions.tex}

\usepackage{todonotes}

\title{
UnitedQA: A Hybrid Approach for Open Domain Question Answering
}

\author{
Hao Cheng\textsuperscript{1}\thanks{~~Equal Contribution} , Yelong Shen\textsuperscript{2}$^*$, Xiaodong Liu\textsuperscript{1}, Pengcheng He\textsuperscript{2}, \\
\textbf{Weizhu Chen\textsuperscript{2}, Jianfeng Gao\textsuperscript{1}}
 \\ 
  \textsuperscript{1} Microsoft Research 
  \textsuperscript{2} Microsoft Azure AI

 \\
  {\tt \{chehao,yeshe,xiaodl,penhe,wzchen,jfgao\}@microsoft.com}
}

\date{}

\aclfinalcopy

\begin{document}
\maketitle

\begin{abstract}
To date, most of recent work under the retrieval-reader framework for open-domain QA focuses on either extractive or generative reader exclusively.
In this paper, we study a hybrid approach for leveraging the strengths of both models.
We apply novel techniques to enhance both extractive and generative readers built upon recent pretrained neural language models,
and find that proper training methods can provide large improvements over previous state-of-the-art models. 
We demonstrate that an hybrid approach by combining answers from both readers can effectively take advantages of extractive and generative answer inference strategies and
outperform single models as well as homogeneous ensembles.
Our approach outperforms previous state-of-the-art models by $3.3$ and $2.7$ points in exact match on NaturalQuestions and TriviaQA respectively.

\end{abstract}

\section{Introduction}
\label{sec:intro}
Open-domain question answering (QA) has been a long standing problem in natural language understanding, information retrieval, and related fields \cite{chen2020open}.
An typical open-domain QA system follows the retrieval-reader framework \cite{drqa,guu2020realm,karpukhin-etal-2020-dense}, where the relevant passages are first retrieved from a large text corpus, and a reader module then navigates multiple passages for answer inference. 
In this work, we study two paradigms of reader modules, \ie \textit{extractive} \cite{karpukhin-etal-2020-dense,guu2020realm} and \textit{generative} \cite{lewis2020retrievalaugmented,izacard2020leveraging} readers. The extractive reader extracts contiguous spans from the retrieved passages whereas the
generative reader sequentially decodes the answer string which might not be contained in the retrieved passages.

\begin{figure}[t]
    \centering
    \includegraphics[width=0.48\textwidth]{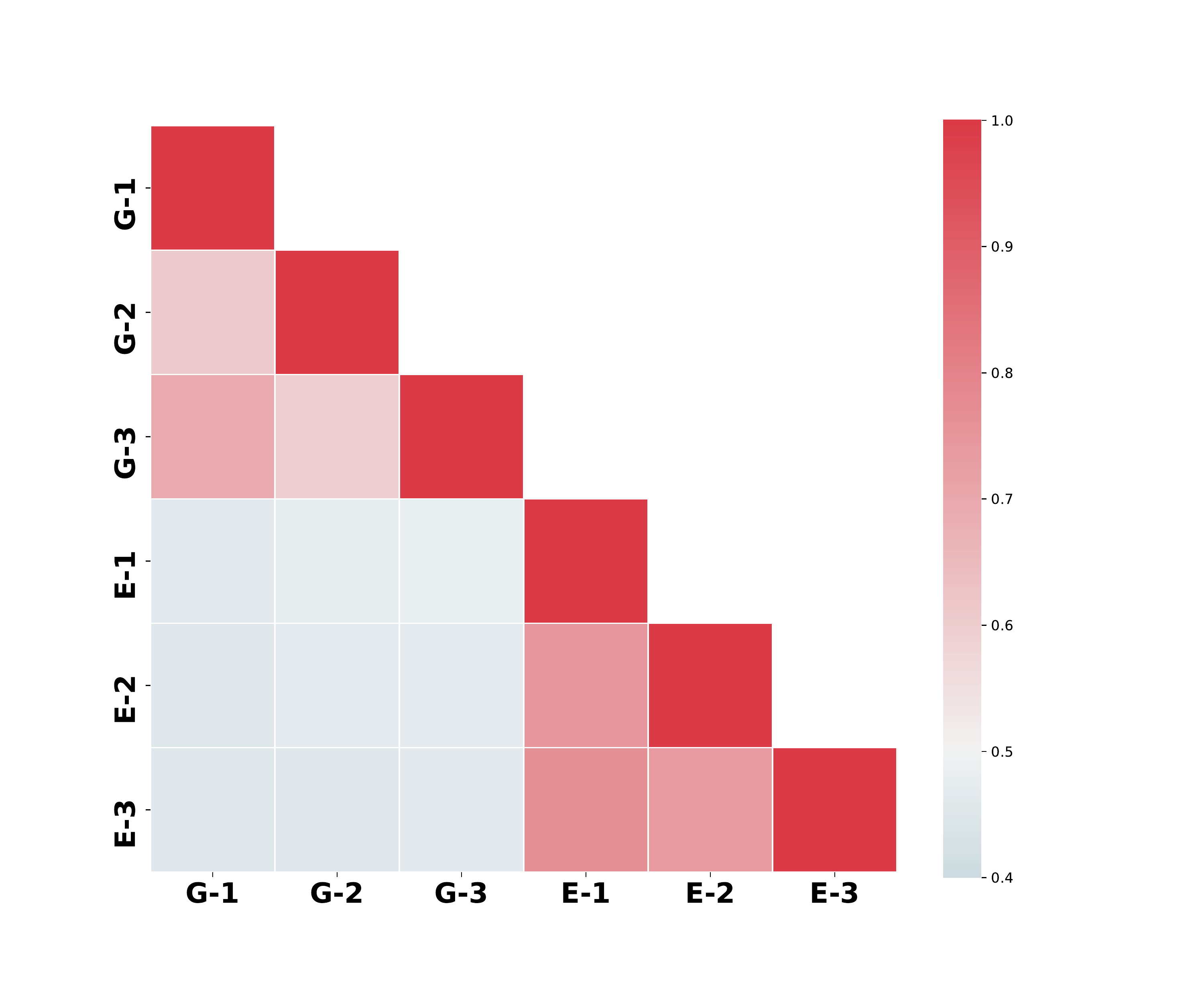}
    \caption{Pairwise prediction agreement ratio. \texttt{G-1, G-2, G-3} and \texttt{E-1, E-2, E-3} are three different generative and extractive readers respectively. All readers achieve similar performance ($\approx52\%$ exact match) on NaturalQuestions. Higher agreement (>$50\%$) in \texttt{red} and lower agreement (<$50\%$) in \texttt{gray}. The agreement is calculated based on exact string match.}
    \label{fig:pred_agree_ratio}
\end{figure}

Recent work on open-domain QA \cite{karpukhin-etal-2020-dense,guu2020realm, lewis2020retrievalaugmented,izacard2020leveraging} explores either an extractive reader or a generative reader exclusively. We hypothesize that extractive and generative readers adopt different answer inference strategies, thus a hybrid extractive/generative reader can be a better option for open-domain QA tasks.
As shown in \autoref{fig:pred_agree_ratio}, compared with prediction agreement among only generative or extractive readers (top-left and bottom-right),
the cross prediction agreement between extractive and generative readers (bottom-left) is relatively low (<$50\%$).
It indicates that answers produced by those two types of models are different and they can be complementary to each other.
Therefore, we propose a hybrid reader approach, UnitedQA, which is a simple ensemble approach to combine the predictions from extractive and generative readers.
It achieves state-of-the-art results on NaturalQuestions \cite{nq} and TriviaQA \cite{joshi-etal-2017-triviaqa}.

In UnitedQA, the extractive reader (UnitedQA-E) and generative reader (UnitedQA-G)
are built upon the pretrained language models, ELECTRA \cite{clark2020electra} and T5 \cite{raffel2020exploring}, respectively.
For the UnitedQA-E, we adopt a weakly-supervised training objective to address the noisy supervision issue caused by the heuristics-based labeling and incorporate the posterior differential regularization (PDR) \cite{cheng2020posterior} to improve the model robustness.
The UnitedQA-G follows the T5 Fusion-in-Decoder (FID) \cite{izacard2020leveraging} and we make two improvements:
first, we add a group of attention bias parameters into the decoder cross-attention block to feature the ranking information of retrieved contexts; second, we add the adversarial training \cite{ju2019technical,jiang2019smart,pereira-etal-2021-targeted} to improve the model generalization ability.

The experimental results highlight the effectiveness of the simple hybrid approach of UnitedQA.
With both improved extractive and generative readers, UnitedQA sets new state-of-the-art results on two popular open-domain QA datasets, \ie $54.7$ and $70.3$ in exact match on NaturalQuestions \cite{nq} and TriviaQA \cite{joshi-etal-2017-triviaqa}, respectively.
It is worth noting that our UnitedQA model not only outperforms each single model but also brings more pronounced improvements over homogeneous ensembles of either extractive or generative readers.
Last, based on our analyses, UnitedQA-E and UnitedQA-G have advantages in different cases, suggesting they may use different reasoning strategies.

\section{Method}
\label{sec:system_overview}
In this section, we present the overall pipeline of the UnitedQA system, which consists of  three components: \textbf{Retrieval}, \textbf{Reading}, and  \textbf{Re-ranking}. 
First, the retrieval module  fetches a list of relevant passages from a Wikipedia dump for a given question. 
Then, the module of hybrid readers produces answer candidates from the set of retrieved passages.
Last, the re-ranking module combines the answer candidates with linear interpolation and produce the final answer.

\noindent
\textbf{Retrieval} Following \citet{karpukhin-etal-2020-dense}, we consider two methods, BM25 and dense passage retrieval (DPR), for retrieving the support passages for a given question.
For BM25, passages are encoded as bag of words (BOW), and inverse document frequencies are used as the ranking function.
For DPR, passages and questions are represented as dense vectors based on two BERT \cite{devlin-etal-2019-bert} models. The relevance score is then computed based on the dot production between the query and passage vectors.
In this paper, we adopt the same implementation as \citet{karpukhin-etal-2020-dense} for retrieving passages. 
Specifically, the English Wikipedia dump from Dec. 20, 2018 is used as the source documents for retrieval, with the removal of semi-structured data, such as tables or lists.
Each document is split into disjoint 100-word \textit{passages} as the basic retrieval unit. The top-100 passages are then passed for reading.

\noindent
\textbf{Reading} We combine the generative reader and the extractive reader to produce answer candidates over the retrieved passages.
Here, we only give a high-level description of our approach.
More details regarding our improved extractive and generative models are presented in \S\ref{sec:united_qa_extractive} and \S\ref{sec:united_qa_generative} respectively.

The generative reader is based on a sequence-to-sequence model pre-trained in a forward-generation fashion on a large corpus, \ie T5 \cite{raffel2020exploring}.
Similar to \citet{izacard2020leveraging}, the model takes the question and its relevant passages as input, and then generates the answer string token by token.
Specifically, the concatenation of all retrieved passages and the corresponding question is used as the encoder input.
Then, the decoder performs reasoning over the concatenation of all evidence through an attention mechanism.

Following state-of-the-art extractive QA models \cite{devlin-etal-2019-bert,karpukhin-etal-2020-dense}, our extractive reader is based on a Transformer neural network pre-trained with a cloze style self-supervised objective, \ie ELECTRA \cite{clark2020electra}. Here, a pair of a given question and a support passage is jointly encoded into neural text representations.
These representations are then used to define scores or probabilities of possible answer begin and end positions, which are in turn used to define probabilities over possible answer spans. Finally, the answer string probabilities are based on the aggregation over all possible answer spans from the entire set of support passages.

\subsection{UnitedQA-E}
\label{sec:united_qa_extractive}
In \S\ref{ssec:improve_extractive}, we give the problem definition of open-domain QA for extractive reader. Then, we detail the improvements of UnitedQA-E in \S\ref{ssec:improve_extractive}. 


\subsubsection{Extractive Reader}
\label{ssec:extractive_reader}
Given a question $\text{q}$ and a set of $K$ retrieved passages $\text{p}_1, \ldots, \text{p}_K$,
a text encoder produces contextualized representations: $\hvec^k_1,...\hvec^k_T \in\RR^n$ for the question-passage pair ($\text{q}, \text{p}_k$) in the form of ``\clstoken \textit{question} \septoken \textit{passage} \septoken'', where \clstoken and \septoken are special tokens for encoding inputs, $T$ is the maximum sequence length of the input text, and $\hvec_i^k$ indicates the contextualized embedding of the $i$-th token in  ($\text{q}, \text{p}_k$). 

The extractive reader computes the span-begin score of the $i$-th token as 
$s_b(i^k) = \wvec_b^T \hvec_i^k$ using a weight vector
$\wvec_b \in \RR^d$.
The span-end score $s_e(j^k)$ is defined in the same way.
Thus, the probabilities of a start position $i^k$ and an end position $j^k$ are 
$  P_b(i^k) = {\exp(s_b(i^k)) \over Z_b},
    P_e(j^k) = {\exp(s_e(j^k)) \over Z_e}, $
where $Z_b, Z_e$ are normalizing factors defined by the corresponding probability space.
The probability of an answer span from $i^k$ to $j^k$ is defined as $P_s(i^k, j^k)=P_b(i^k)P_e(j^k)$.

Here, we consider two probability spaces, \textbf{passage level} and \textbf{multi-passage level}, with the only difference in the computing of $Z_b, Z_e$.
Specifically, the passage-level probability of each answer begin and end is computed by normalizing all possible positions in the respective passage, \ie $Z_b = Z_b^k = \sum_{\mathcal{I}^k \cup \emptyspan}\exp(s_b(i)), Z_e=Z_e^k=\sum_{\mathcal{I}^k \cup \emptyspan}\exp(s_e(j))$, where $\mathcal{I}^k$ is the set of all possible positions from the $k$-th passage and \emptyspan\  indicates special positions if $p_k$ does not support answering the question.
Similarly, the multi-passage level probability is computed by normalizing over each answer positions across all $K$ relevant passages, \ie
$Z_b = Z_b^* = \sum_k \sum_{\mathcal{I}^k} \exp (s_b(i)), Z_e = Z_e^* = \sum_k \sum_{\mathcal{I}^k} \exp (s_e(j))$, respectively.

Since there are usually multiple plausible mentions for open-domain QA, during training, it is typical to maximize either the marginal log-likelihood (MML) of all correct spans \cite{karpukhin-etal-2020-dense} or the log-likelihood of the most likely correct span (HardEM) \cite{min-etal-2019-discrete}.
During inference, the prediction is made based on the candidate answer string score, obtaining as $P_a(y)=\sum_{(i, j)\in\mathcal{Y}}P_s(i, j)$, where
$\mathcal{Y}$ is the set of spans corresponding to the answer string $y$.

\subsubsection{Improvement Method}
\label{ssec:improve_extractive}
In addition to better text representations from \citet{clark2020electra}, we consider two methods for improving the training of the extractive reader.

\noindent
\textbf{Multi-objective for Weakly-supervised QA} The multi-objective formulation is introduced in \citet{cheng-etal-2020-probabilistic} for improving weakly supervised document-level QA.
Different from \citet{cheng-etal-2020-probabilistic} where only MML is considered for the multi-objective formulation, we found combining HardEM with MML is more effective for open-domain QA based on our experiments (\S\ref{ssec:ablative_study_ext}). Specifically, we combine a multi-passage HardEM loss with $K$ passage-level MML losses over a batch of $K$ passages
\begin{eqnarray}
\nonumber
    \mathcal{L}_\text{EXT} & = & \log \max_{(i, j)} P^M_s(i, j) + \\
    & & {1\over K} \sum_k \log \sum_{(i^k, j^k)} P^P_s(i^k, j^k),
\end{eqnarray}
where $P^M_s$, $P^P_s$ is the multi-passage level and passage level span probabilities respectively.
    
\noindent
\textbf{Posterior Differential Regularization} 
Due to the noisy supervision for open-domain QA \cite{drqa}, we investigate the posterior differential regularization (PDR) \cite{cheng2020posterior} to improve the robustness of the extractive reader.
Different from \citet{cheng2020posterior} where only clean supervision setting is considered, in this work, we apply PDR to the weakly supervised open-domain QA scenario. 
Given it is computationally expensive to enumerate all possible spans, we apply two separate regularization terms for the begin and end probabilities at the multi-passage level, respectively,
\begin{eqnarray}
   \mathcal{L}_\text{PDR} = D(P_b(i) | P^\prime_b(i) ) + D(P_e(j) | P^\prime_e(j)),
\end{eqnarray}
where $D(\cdot|\cdot)$ is the squared Hellinger distance, and $P^\prime_b, P^\prime_e$ are the probabilities of start and end positions with additive input noise to the token embeddings. Specifically, we sample noise vectors $\epsilon_1, \ldots, \epsilon_T$ from $\mathcal{N}(0, c^2I)$,
and add them to the token embeddings as the noisy input, \ie $\vvec_1+\epsilon_1, \ldots, \vvec_T+\epsilon_T$, where $c$ is fixed to $1\mathrm{e}{-3}$ throughout our experiments.

Based on this, the overall training objective for the extractive reader is 
\begin{eqnarray}
   \mathcal{L}^1 = \mathcal{L}_\text{EXT} + \gamma \mathcal{L}_\text{PDR},
\end{eqnarray}
where $\gamma$ is a regularization scalar hyperparameter.

\subsection{UnitedQA-G}
\label{sec:united_qa_generative}
Here, we first formally define the setup of generative reader for open-domain QA in \S~\ref{ssec:generative_reader} and then present our improvements in \S~\ref{ssec:improve_generative}.


\subsubsection{Generative Reader}
\label{ssec:generative_reader}
Given a question $\text{q}$ and a set of $K$ retrieved passages $\text{p}_1, \ldots, \text{p}_K$, the encoder model encodes each $(\text{q}, \text{p}_k)$ pair independently, and produces contextualized representation for each token: $\hvec_i^k \in \RR^d$ for the $i$-th token of the $k$-th pair. The decoder then performs attention over the concatenation of the representations of all the retrieved passages, and generates the answer string. 

Let $\mathbf{x}$ denote the input of the question and all retrieved passages $\mathbf{x} = \big( (\text{q}, \text{p}_1), ..., (\text{q}, \text{p}_K) \big) $, and $\mathbf{y}$ the answer string with its tokens as $(y_1, ..., y_N)$.
The generative reader is trained to maximize a sequence-to-sequence objective for a given $(\mathbf{x}, \mathbf{y})$, 
\begin{align}
\label{eq:generative_obj}
 \mathcal{L}(\xvec, \yvec;\theta) = \sum_{i}^{N} \text{log} P_{\theta}(y_i | \mathbf{x}, y_{1:i-1}),
\end{align}
where $\theta$ is the model parameter.
During inference, a greedy decoding is used to produce the answer.

\subsubsection{Improvement Method}
\label{ssec:improve_generative}
\noindent
\textbf{Decoder Attention Bias} The decoder in the T5 transformer model adopts a cross-attention mechanism to compute attention scores between the decoding answer tokens and all the retrieved passage tokens. Specifically, let $\yvec_i \in \RR^d$ be the query vector of the $i$-th decoding token\footnote{we omit the layer notation for simplification},  and $\mvec^k_j \in \RR^d$ be the key vector of the $j$-th token in $(\text(q), \text{p}_k)$. The multi-head cross-attention scores in T5 \cite{raffel2020exploring} $\vec{s}^k_{i,j}$ is calculated as
\begin{align}
\label{eqn:mh_standard}
\svec^k_{i,j} = \text{MultiHeadAtt}( \yvec_i , \mvec^k_{j}) \in \RR^{|\text{Head}|}
\end{align}
where $|\text{Head}|$ is the number of attention heads.
However, it doesn't capture the relevance information of retrieved passages into the reader in \eqref{eqn:mh_standard}. To add the relevance feature into the attention block, 
we revise \eqref{eqn:mh_standard} by incorporating the attention bias 
\begin{align}
\label{eqn:mh_standard_bias}
\svec^k_{i,j} = \text{MultiHeadAtt}( \yvec_i , \mvec^k_{j}) +\bvec_{k},
\end{align}
where $\bvec_k \in \RR^{|\text{Head}|}$ is a trainable attention bias vector for all the tokens in the $k$-th retrieved passage. In the experiments, the maximum retrieved passages is by default set to $100$. Thus, the decoder attention bias introduces additional $100 * |\text{Head}|$ parameters for each layer.


\noindent\textbf{Adversarial Training}
Adversarial training creates \emph{adversarial} examples by adding small perturbations to the embedding layer. Assuming the word(-piece) embedding layer is parameterized by a matrix $\mathbf{V} \in \mathcal{R}^{|V| \times d} $, $|V|$ is the vocabulary size, and $d$ is the embed-dimension. The adversarial embedding matrix $\hat{\mathbf{V}}$ can be obtained by  
\begin{align}
g_{\mathbf{V}} = - \nabla_{\mathbf{V}} \mathcal{L}(\mathbf{x}, \mathbf{y}; \theta), \\
\hat{\mathbf{V}} = \mathbf{V} + \text{SG} ( \epsilon g_{\mathbf{V}} / || g_{\mathbf{V}} ||_2 ),
\end{align}
where $\text{SG}(\cdot)$ is the stop-gradient operation.
We use the adversarial embedding matrix $\hat{\mathbf{V}}$ to replace the original $\mathbf{V}$ in model parameters $\theta$, and obtain $\hat{\theta}$. 
Thus the adversarial loss can be calculated as  
\begin{align}
\mathcal{L}_{\text{AT}}(\mathbf{x}, \mathbf{y}; \theta) = \mathcal{L}_{}(\mathbf{x}, \mathbf{y}; \hat{\theta}).
\end{align}

Therefore, the overall training objective of the generative reader is
\begin{align}
\mathcal{L}^2 = \alpha \mathcal{L}(\mathbf{x}, \mathbf{y}; \theta) + \beta    \mathcal{L}_{\text{AT}}(\mathbf{x}, \mathbf{y}; \theta),
\end{align}
where $\alpha=0.5, \beta=0.5$ in all of the exepriments.

\subsection{UnitedQA System}
The UnitedQA system combines outputs from both extractive and generative models for a given question during inference. 
Since the output spaces of extractive and generative models are different,
we use a simple linear interpolation based on best predictions from each model\footnote{%
We have also tried a few more complex approaches for combining the extractive and generative models.
For example, we first train an extractive model, and then append the top-k answer strings from the extractive model at the end of the input for training a generative model. 
None of them is as good as the simple ensemble approach.
}.
Denote the predicted strings from $M$ extractive and $N$ generative models as $y^{E}_{1}, ..., y^{E}_{M}$
and $y^{G}_{1}, ..., y^{G}_{N}$, respectively.
The hybrid prediction $y^*$ is obtained by
\begin{align}
    \argmax_{y \in \mathcal{Y}} \tau \sum_{m=1}^{M} \one(y, y^E_{m}) +  \delta \sum_{n=1}^{N} \one(y, y^G_{n}),
    \label{eqn:hybrid_prediction}
\end{align}
where $\mathcal{Y}$ is the set of all predicted strings, $\one(y, y^\prime)$ is an indicator function and $\tau=0.6, \delta=0.4$. 


\section{Experiments}
\label{sec:experiments}
\begin{table}[t]
    \centering
    \begin{tabular}{lccc}
    \hline
\toprule
Dataset &  Train  & Dev & Test\\
\midrule
NQ  & 79168 & 8757 & 3610 \\
TriviaQA & 78785 &  8837 & 11313 \\
EffcientQA & - & 1800 &-\\
\bottomrule
    \end{tabular}
    \caption{Number of questions in each QA dataset.}

    \label{tab:datasets}
\end{table}

\begin{table*}[t]
    \centering
    \begin{tabular}{l|c|c|c|c}
    \hline
\toprule
Model 
& Reader Type & Reader Size (M) & NQ & TriviaQA \\
\midrule
REALM\cite{guu2020realm}  
& Extractive & 110  &  40.4 & N/A \\
RAG\cite{lewis2020retrievalaugmented}  
& Generative & 400 & 44.5 & 56.1 \\
\midrule
DPR\cite{karpukhin-etal-2020-dense}    
& Extractive & 110 & 41.5 & 57.9 \\
T5-FID\textsubscript{base}\cite{izacard2020leveraging} 
& Generative & 220 & 48.2 & 65.0 \\
T5-FID\textsubscript{large}\cite{izacard2020leveraging} 
& Generative & 770 & 51.4 & 67.6 \\
\midrule
UnitedQA-E\textsubscript{base} (Ours) 
& Extractive & 110 & \underline{47.7} & \underline{66.3} \\
UnitedQA-E\textsubscript{large} (Ours)
& Extractive & 330 & 51.8 & \textbf{68.9}\\
UnitedQA-G\textsubscript{large}(Ours) & Generative & 770 & \textbf{52.3} & 68.6\\
\midrule
UnitedQA-E\textsubscript{large}++ (Ours) & Ensemble & 3x330 & 52.4 & 69.6 \\
UnitedQA-G\textsubscript{large}++ (Ours) & Ensemble & 3x770 & 53.3 & 69.2 \\
UnitedQA (Ours) & Hybrid & 2x770+330 & \colorbox{Gray}{\textbf{54.7}} & \colorbox{Gray}{\textbf{70.5}} \\
\bottomrule
    \end{tabular}
    \caption{Comparison to state-of-the-art models on the test sets of NaturualQuestions (NQ) and TriviaQA. Exact match score is used for evaluation. The overall best model is in \colorbox{Gray}{\textcolor{Gray}{Box}}, the best single model is in \textbf{bold}, and the best model with the smallest reader size is in \underline{underline}.}
    \label{tab:unitedqa_vs_sota}
\end{table*}

\subsection{Experiment Setup}

We use two representative QA datasets and adopt the same training/dev/testing splits as in previous work \cite{lee-etal-2019-latent,karpukhin-etal-2020-dense}.
Both datasets (see \autoref{tab:datasets} for statistics) have been heavily studied in recent work \cite{lee-etal-2019-latent,min-etal-2019-discrete,karpukhin-etal-2020-dense,guu2020realm}.
We follow the standard evaluation protocol and use exact match (EM) as the evaluation metric.

\noindent
\textbf{NaturalQuestions \cite{nq}} is composed of questions by real users to Google Search, each with answers identified by human annotators in Wikipedia.
The open-domain version of NaturalQuestions \cite{lee-etal-2019-latent} only consider questions with short answers, \ie answers with less than 5 tokens. 
In the NaturualQuestions, the questions are considered to be more information seeking given that the question askers didn't know the answer beforehand.
In addition, we use another evaluation set, \ie the dev set introduced recently by the EfficientQA competition \cite{min2021neurips}, which is constructed in the same way as the original NaturalQuestions dataset.

\noindent
\textbf{TriviaQA \cite{joshi-etal-2017-triviaqa}} contains trivia question-answer pairs that were scraped from the web.
Different from NaturalQuestions, the questions here are written with known answers in mind.
Specifically, the unfiltered set has been used for developing open-domain QA models.

\noindent
\textbf{Implementation details} 
For a fair comparison, we use the same retrieval module as \citet{karpukhin-etal-2020-dense} for NaturalQuestions and TriviaQA to mitigate the impact of retrieval difference.
Specifically, we use DPR (single) for NaturalQuestions and BM25+DPR (multi) for TriviaQA because of their best end-to-end performance (Karpukhin et al. 2020).
For all the experiments, we use 8 and 16 V100-32GB for base and large model training respectively.
We train our models with Adam optimizer of a linear scheduler with a warmup raito of 0.1.
The extractive models are trained for up to 8 epochs with a learning rate of $2\mathrm{e}{-5}$ and a batch passage size per question of $16$.
The generative models are trained for up to 10 epochs with a learning rate of $1\mathrm{e}{-4}$, a batch size of $64$, and $100$ retrieved passages per question for model training. 
We select $\gamma$ in $\{4, 8\}$. After the best configuration is selected based on the dev set, we run our best models 3 times independently with different random seeds and report the median performance on the test set. We also report ensemble results which are based on the linear interpolation over answer predictions from the 3 models.
\subsection{Main results}
\label{ssec:exp_ext_results}

\noindent\textbf{Single Model Results:} We first compare our models to two recent models, REALM \cite{guu2020realm} and RAG \cite{lewis2020retrievalaugmented}, which are first pre-trained with different retrieval augmented objectives and then fine-tuned for open-domain QA.
In addition, we include as baselines DPR \cite{karpukhin-etal-2020-dense} and T5-FID \cite{izacard2020leveraging}, both of which are based on the same retriever as ours.
As shown in \autoref{tab:unitedqa_vs_sota}, both our extractive and generative models achieve new state-of-the-art results for both studied datasets.
Compared with the recent state-of-the-art extractive model (DPR), our base model leads to pronounced $15\%$ relative improvements for both NaturalQuestions ($+6.2$ absolute improvement) and TriviaQA ($+8.4$ absolute improvement). More importantly, UnitedQA-E\textsubscript{base} achieves comparable or even better performance with regard to generative models of larger size, \ie RAG and T5-FID\textsubscript{base}. It highlights the importance of proper training strategies for open-domain QA models.

\noindent{\textbf{Hybrid Model Results:} In order to evaluate the advantage of the hybrid of the extractive and generative models (UnitedQA), we include two homogeneous ensemble baselines, one consisting of only extractive readers (UnitedQA-E++) and the other ensemble of exclusively generative models (UnitedQA-G++).
For homogeneous ensemble cases, the three-way majority prediction is used.
For the hybrid of extractive and generative readers, we select a three-model combination from the set of three generative and three extractive models based on the dev set. We observed that combining predictions from two generative models and one extractive model results in the best hybrid model for both datasets.
As expected, all ensemble models show an improvement over their single model counterparts.
However, the two homogeneous ensemble baselines, UnitedQA-E++ and UnitedQA-G++, only provide marginal gains over the corresponding best single models.
The significant improvement brought by our proposed hybrid approach indicates the benefit of combining extractive and generative readers for open-domain QA.

\noindent\textbf{Discussion:} Although the proposed hybrid approach has been shown to be highly effective for open-domain QA, we point out that the improved performance comes with increased computational cost. The best combination requires approximately three times the computational cost of a single generative model. 
Therefore, it would be interesting to explore more efficient hybrid methods, such as effective parameter sharing strategies or unified formulations.
Another interesting future direction is to explore customized compression approaches for reducing the model size of retriever and reader separately or jointly through pruning \cite{han2015pruning}, quantization \cite{hubara2018quantization}, and knowledge distillation \cite{hinton2015kd}.
Specifically, given that the hybrid model is more effective, it is likely that a student model can learn more effectively from a hybrid teacher model via knowledge distillation for open-domain QA.

\section{Analysis}
\label{sec:analysis}
\begin{table}[t!]
    \centering
    \begin{tabular}{l|c|c}
    \hline
\toprule
Model &  NQ & TriviaQA \\
\midrule
\cite{cheng-etal-2020-probabilistic}
+\texttt{PDR} &  43.3 & 60.1 \\
BERT\textsubscript{base}    & 44.2  & 62.2\\
\hspace{0.1in}-\texttt{Multi-obj } & 43.5  &  61.3\\
\hspace{0.1in}-\texttt{PDR} & 41.8  & 60.2 \\
\hspace{0.1in}-\texttt{Multi-obj \& PDR} & 40.6  & 58.5 \\
\midrule
UnitedQA-E\textsubscript{base}    & 46.0  & 65.4\\
\hspace{0.1in}-\texttt{Multi-obj } & 45.2  &  64.3\\
\hspace{0.1in}-\texttt{PDR} & 43.1  &  63.8\\
\hspace{0.1in}-\texttt{Multi-obj \& PDR} & 42.5  & 61.2 \\
\bottomrule
    \end{tabular}
    \caption{Ablation experiments of the extractive model on the dev sets of NaturalQuestions (NQ) and TriviaQA. Exact match score is reported. The top and bottom models are built on BERT\textsubscript{base} and ELECTRA\textsubscript{base}, respectively.}
    \label{tab:extractive_ablation}
\end{table}
In this section, we first carry out ablation study on the extractive and generative model improvements.
Moreover, we aim to take a deeper look and understand the difference between the two models.

\subsection{Ablation Study}
\label{ssec:ablative_study_ext}
In \autoref{tab:extractive_ablation}, we present ablation experiments on the effectiveness of different textual representations and methods for improving the extractive model UnitedQA-E\textsubscript{base}. Here, we focus on base models, \ie BERT\textsubscript{base} and ELECTRA\textsubscript{base}. Note that the row \texttt{UnitedQA-E\textsubscript{base}} is the corresponding base model reported in \autoref{tab:unitedqa_vs_sota}.
Compared with the MML-based multi-objective \cite{cheng-etal-2020-probabilistic}, we find that a new multi-objective with HardEM at the multi-passage level and MML at the passage level is more effective for open-domain QA.
In addition to the multi-objective training, there is a noticeable improvement brought by the regularization method (PDR) which indicates the importance of proper regularization for learning with noisy supervision.
Last but not least, the large improvement of ELECTRA over BERT indicates the importance of deriving better text representations for weakly supervised NLP problems.
For the UnitedQA-G, we present the ablation study on analyzing the effectiveness of decoder attention bias component and adversarial training mechanism in \autoref{tab:generative_ablation}. Both techniques contribute to decent improvements over T5-FID with more pronounced gains brought by adversarial training.

\begin{table}[t]
\centering
\begin{tabular}{l|c|c}
\hline
\toprule
Model &  NQ & TriviaQA \\
\midrule
T5-FID\textsubscript{large} &  51.4 & 67.6 \\
UnitiedQA-G\textsubscript{large}  &  52.3 & 68.6 \\
\hspace{0.1in}-\texttt{Adv Training} & 52.0  & 68.2 \\
\hspace{0.1in}-\texttt{Attention Bias} & 51.8  & 68.1 \\
\bottomrule
\end{tabular}
\caption{Ablation experiments of the generative model on the test sets of NaturalQuestions (NQ) and TriviaQA. Exact match score is reported.}
\label{tab:generative_ablation}
\end{table}

\subsection{Impact of Retrieval Accuracy} 
Here, we vary the number of retrieved passages during inference and report the evaluation results in terms of end-to-end QA exact match score of UnitedQA-E and UnitedQA-G along with the corresponding top-$k$ retrieval accuracy.
The results are summarized in \autoref{tab:topk_eval}.
As expected, when the number of retrieved passages increases, both top-$k$ retrieval accuracy and the end-to-end QA performance improve.
However, there is a noticeable gap between the improvement of retrieving more passages (i.e., recall) and that of the corresponding end-to-end QA performance, especially for the extractive reader.
This is likely caused by additional noise introduced with improved retrieval recall.
Specifically, only half of the retriever improvement can be effectively utilized by the extractive model while the generative model can benefit more from retrieving more passages.
This suggests that by concatenating all passages in vector space, the generative model are more effective in de-noising in comparison to the extractive model.

\begin{table}[t]
    \centering
    \begin{tabular}{@{\hskip3pt}c@{\hskip3pt}|c|c|c|@{\hskip3pt}c@{\hskip3pt}}
    \hline
\toprule
& & Top-20  & Top-100 & $\Delta$ \\
\midrule
\multirow{3}{*}{NQ}
& Retrieval & 78.4 & 85.4 & +9$\%$ \\
& United-E & 49.8 & 51.8 & +4$\%$ \\
& United-G & 49.3 & 52.3 & +6$\%$ \\
\midrule
\multirow{3}{*}{TriviaQA}
& Retrieval & 79.9 & 84.4 & +6$\%$ \\
& United-E & 67.1 & 68.9 & +3$\%$ \\
& United-G & 65.4 & 68.6 & +5$\%$ \\
\bottomrule
    \end{tabular}
    \caption{Retieval top-$k$ accuracy and end-to-end QA extact match scores on the test sets of NaturalQuestions (NQ) and TriviaQA. United-E and United-G stand for our extractive and generative models respectively.}
    \label{tab:topk_eval}
\end{table}

\begin{table*}[t]
    \centering
     \begin{tabular}{@{\hskip2pt}cc|c|M{1.5cm}M{1.5cm}M{1.5cm}M{1.5cm}M{1.5cm}@{\hskip2pt}}
    \hline
\toprule
  Dataset & Model & Total & Question Overlap & No Question Overlap & Answer Overlap &Answer Overlap Only & No Overlap \\
\midrule
\multirow{2}{*}{NQ}
& UnitedQA-G     & \textbf{52.3} & \textbf{72.2} & 40.5 & \textbf{62.7} & \textbf{45.4} & 34.0 \\
& UnitedQA-E     & 51.8          & 69.4 & \textbf{41.5} & 60.1 & 45.1 & \textbf{37.6} \\
\midrule
\multirow{2}{*}{TriviaQA}
& UnitedQA-G     & 68.6          & 88.4          & 62.5          & 78.1          & 69.6          & \textbf{44.5} \\
& UnitedQA-E     & \textbf{68.9} & \textbf{89.3} & \textbf{62.7} & \textbf{78.6} & \textbf{70.6} & 44.3 \\
\bottomrule
    \end{tabular}
    \caption{Breakdown evaluation on NaturalQuestions (NQ) and TriviaQA based on test splits defined in \cite{lewis2020question}. Exact match scores are reported. UnitedQA-E and UnitedQA-G denote our extractive and generative models respectively.}
    \label{tab:breakdown_eval}
\end{table*}

\subsection{Breakdown Evaluation}
Following \citet{lewis2020question}, we carry out a breakdown evaluation of model performance over the NaturalQuestions and TriviaQA test sets.
Given their superior performance, we again only consider our improved extractive and generative models, \ie UnitedQA-E\textsubscript{large} and UnitedQA-G respectively.
The evaluation is summarized in \autoref{tab:breakdown_eval}.
In comparison to their corresponding overall performance, both the extractive and generative models achieve much better performance on the ``Overlap'' categories (\ie ``Question Overlap'' and ``Answer Overlap'') for both NaturalQuestions and TrivaQA, which indicates that both models perform well for question and answer memorization.
Different from question and answer memorization, there is a pronounced performance drop for both models on the``Answer Overlap Only'' category where certain amount of relevance inference capability is required to succeed.
Lastly, we see that both extractive and generative models suffer some significant performance degradation 
for the ``No Overlap'' column which highlights model's generalization evaluation.
Nevertheless, the extractive model demonstrate a better QA generalization by achieving a better overall performance on the ``No Overlap'' category for both datasets.

\subsection{Error Analysis}
Here, we conduct analyses into prediction errors made by the extractive and generative models based on automatic evaluation.
For this study, we use the EfficientQA dev set \cite{min2021neurips} which is constructed in the same way as the original NaturalQuestions dataset. 
Specifically, we group prediction errors into three categorizes: 1) common prediction errors made by 
both the extractive and generative models, 
2) prediction errors made by the extractive model, 3) prediction errors produced by the generative model.
In the following, we first carry out a manual inspection into the common errors. Then, we compare the prediction errors made by extractive and generative models, respectively.
\begin{table*}[t!]
    \centering
     \begin{tabular}{@{\hskip2pt}c@{\hskip2pt}|@{\hskip2pt}l@{\hskip2pt}}
    \hline
\toprule
\multicolumn{2}{c}{\bf Valid Answers} \\
\midrule
\multirow{2}{*}{Different granularity} & Q: When was harry potter and the deathly hallows part 2 movie released\\
& Prediction: 2011 / Gold: 15 July 2011 \\
\multirow{2}{*}{Semantically equivalent} & Q: minimum age limit for chief justic of india\\
& Prediction: 65 / Gold: 65 years \\
\multirow{2}{*}{Ambiguity question} & Q: who won her first tennis grand slam in 2018\\
& Prediction: Carolin Wozniacki / Gold: Simona Halep\\
\midrule
\multicolumn{2}{c}{\bf Wrong Answers}\\
\midrule
\multirow{2}{*}{Part as whole error} & Q: the official U.S. poverty line is based on the cost of what\\
& Prediction: food / Gold: ICP purchasing power \\
\multirow{2}{*}{Entity confusion} & Q: actor who played tommy in terms of endearment\\
& Prediction: Jeff Daniels / Gold: Troy Bishop\\
\multirow{2}{*}{Event confusion}  & Q: when did the saskatchewan roughriders last won the grey cup\\
& Prediction: 2007 / Gold: 2013\\
\bottomrule
    \end{tabular}
    \caption{Examples of prediction errors as judged by the automatic evaluation.}
    \label{tab:pred_err}
\end{table*}

First of all, there is an error rate of $29\%$ of those consensus predictions made by both extractive and generative models according to the automatic evaluation. Based on $30$ randomly selected examples, we find that around $30\%$ of those predictions are actually valid answers as shown in the top part of \autoref{tab:pred_err}. In addition to predictions that are answers at different granularity or semantically equivalent ones, some of those prediction errors are likely caused by the ambiguity in questions. As the given example in \autoref{tab:pred_err}, based on the specificity, the model prediction is also a valid answer.
This highlights the limitation of the current evaluation metric, which does not accurately estimate the existing open-domain QA system capabilities.
As shown in the bottom part of \autoref{tab:pred_err}, most of representative errors are due to the confusion of related concepts, entities or events that are mentioned frequently together with the corresponding gold answers. 

Next, all questions from the dev set are categorized based the \textit{WH} question word, \ie \textit{what, which, when, who, how, where}.
We then report the relative performance change of each \textit{WH} category for both extractive and generative models over their corresponding overall prediction accuracy in \autoref{fig:rel_acc_wh_type}.
First, it is easy to see that both extractive and generative models achieve the best performance for entity related \textit{who} questions, which is likely to be the result of high ratio of samples of this type seen during training. 
In contrast, the answers to \textit{what} questions can play a much richer syntactic role in context, making it more difficult for both extractive and generative models to perform well.
Interestingly, the generative model exhibits the strength for temporal reasoning, whereas the extractive model does not.
This difference suggests that it is worth exploring better temporal modeling strategies to improve the extractive model in the future.

\begin{figure}
    \centering
    \includegraphics[width=0.48\textwidth]{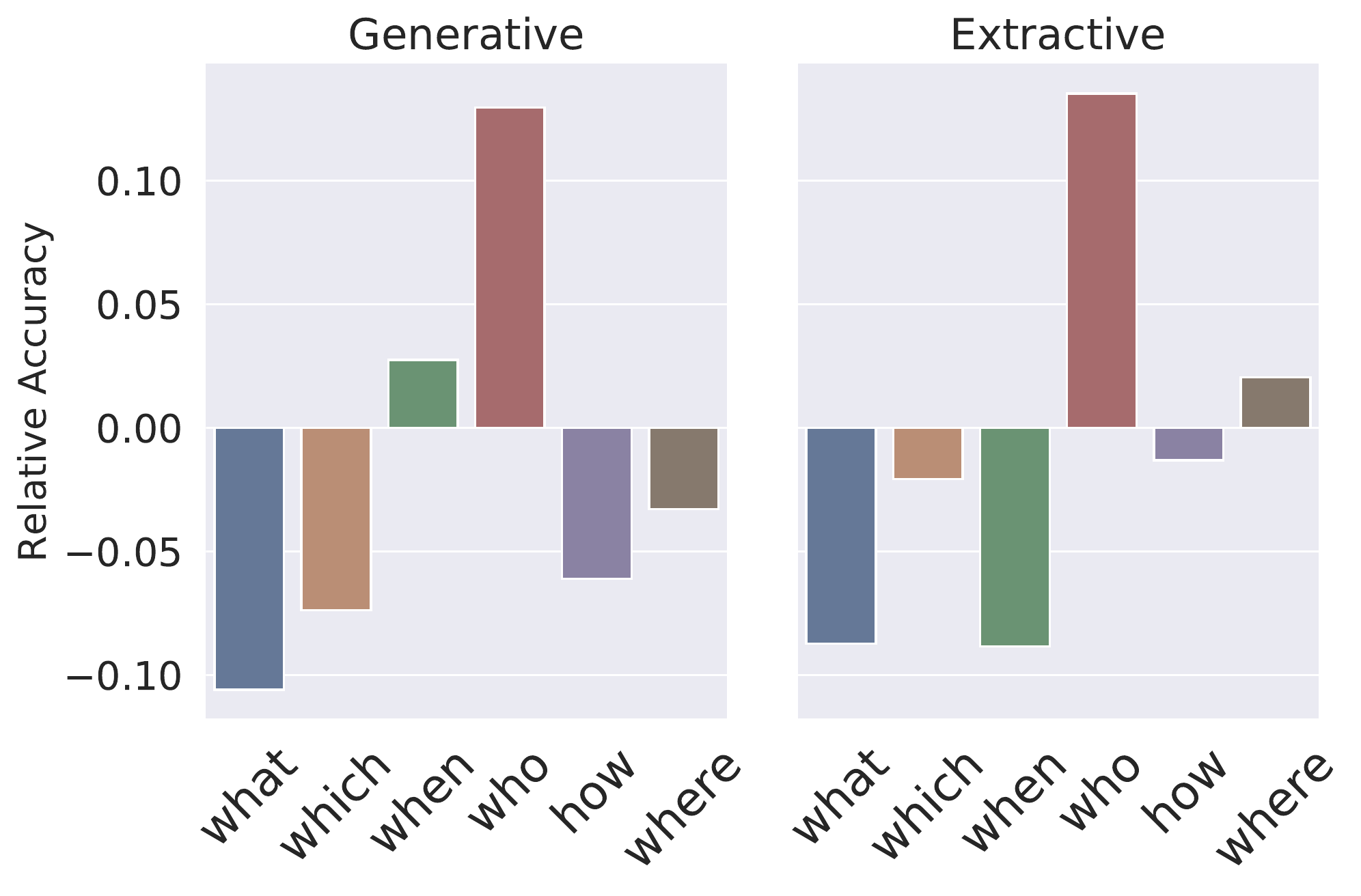}
    \caption{Relative accuracy of different \textit{WH} questions. The relative accuracy is the relative change of a \textit{WH} category accuracy to the overall model accuracy.}
    \label{fig:rel_acc_wh_type}
\end{figure}

\section{Related Work}
\label{sec:related_work}
\textbf{Open-domain QA}
Open-domain QA requires a system to answer questions based on evidence retrieved from a large corpus such as Wikipedia \cite{trec8_qa,drqa}.
Recent progress has been made towards improving evidence retrieval through both sparse vector models like TF-IDF or BM25 \cite{drqa,min-etal-2019-discrete},
and dense vector models based on BERT \cite{lee-etal-2019-latent,karpukhin-etal-2020-dense,guu2020realm,ding2020rocketqa}.
Generally, the dense representations complement the sparse vector methods for passage retrieval
as they can potentially give high similarity to semantically related text pairs, even without exact lexical overlap.
Unlike most work focusing on a pipeline model, \citet{lee-etal-2019-latent} propose a pre-training objective for jointly training both the retrieval encoder and reader.
It is further extended by \citet{guu2020realm} with a dynamic update of the passage index during the training.
Instead, in this work, we focus on a hybrid reader approach for open-domain QA.
By simply combing answer predictions from extractive and generative models, our UnitedQA achieves significant improvements over state-of-the-art models.

\noindent
\textbf{Reading Comprehension with Noisy Labels}
There has been a line of work on improving distantly-supervised reading comprehension models by developing learning methods and model architectures that can better use noisy labels. Most of them focus on the document-level QA, where all paragraphs share the same document context.
\citet{clarkgardner} propose a paragraph-pair ranking objective for learning with multiple paragraphs so that the model can distinguish relevant paragraphs from irrelevant ones.
In \cite{lin-etal-2018-denoising}, a coarse-to-fine model is proposed to handle label noise by aggregating information from relevant paragraphs and then extracting answers from selected ones.
\citet{min-etal-2019-discrete} propose a hard EM learning scheme where only passage-level loss is considered for document-level QA. More recently, different probabilistic assumptions with corresponding training and inference methods are examined in \cite{cheng-etal-2020-probabilistic} again for document-level QA with distant supervision.
In our work, we further extend the multi-objective formulation proposed in \cite{cheng-etal-2020-probabilistic} with the hard EM learning \cite{min-etal-2019-discrete} for
enhancing extractive open-domain QA, where the input passages are given by a retrieval model and are typically from different documents.


\section{Conclusion}
\label{sec:conclusion}
In this study, we propose a hybrid model for open-domain QA, called UnitedQA, which combines the strengths of extractive and generative readers.
We demonstrate the effectiveness of UnitedQA on two popular open-domain QA benchmarks, NaturalQuestions and TriviaQA.
Our results show that the proposed UnitedQA model
significantly outperforms single extractive and generative models as well as their corresponding homogeneous ensembles, and sets new state-of-the-art on both benchmarks.
We also perform a comprehensive empirical study to investigate the relative contributions of different components of our model and the techniques we use to improve the readers. 

For future work, it would be interesting to explore model compression approaches for reducing the model size of retriever and reader separately or jointly through pruning, quantization, and knowledge distillation.
\section*{Acknowledgments}
We would like to thank the anonymous reviewers for valuable suggestions, Yuning Mao for valuable discussions and comments, and Microsoft Research Technology Engineering team for computing support.

\bibliography{ref,qa}
\bibliographystyle{acl_natbib}
\clearpage

\appendix
\setcounter{table}{0}
\renewcommand{\thetable}{A\arabic{table}}

\setcounter{figure}{0}
\renewcommand{\thefigure}{A\arabic{figure}}

\end{document}